\documentclass[sigconf]{acmart}

\usepackage{multirow}
\usepackage{booktabs}
\usepackage{algorithm}
\usepackage{algorithmic}
\usepackage{graphicx}
\usepackage{arydshln} 
\usepackage{amsmath}
\usepackage{amsthm}

\usepackage{subfloat}
\usepackage{subfig}

\usepackage{fancyhdr}
\AtBeginDocument{%
	}

\setcopyright{acmcopyright}
\copyrightyear{2018}
\acmYear{2018}
\acmDOI{XXXXXXX.XXXXXXX}

\acmConference[MM '24]{Proceedings of the 31th ACM International Conference on Multimedia}{October 21--November 01,
	2024}{Melbourne, Australia}

\acmPrice{15.00}
\acmISBN{978-1-4503-XXXX-X/18/06}

\renewcommand\footnotetextcopyrightpermission[1]{}
\begin{document}
	
	\title{Revisiting Multi-modal Emotion Learning with Broad  State Space Models and Probability-guidance Fusion}

	\author{Yuntao Shou}
	\affiliation{
		  \institution{Central South University of Forestry and Technology}
		  \city{Hunan}
		  \country{China}}
	\email{shouyuntao@stu.xjtu.edu.cn}
	
		\author{Tao Meng}
				\authornote{Corresponding Author}
	\affiliation{
		\institution{Central South University of Forestry and Technology}
		\city{Hunan}
		\country{China}}
	\email{mengtao@hnu.edu.cn}

		\author{Fuchen Zhang}
	\affiliation{
		\institution{Central South University of Forestry and Technology}
		\city{Hunan}
		\country{China}}
	\email{fuchen.zhang@csuft.edu.cn}
	
		\author{Nan Yin}
	\affiliation{
		\institution{Mohamed bin Zayed University of Artificial Intelligence
			}
		\country{UAE}}
	\email{nan.yin@mbzuai.ac.ae}
	
			\author{Keqin Li}
	\affiliation{
		\institution{State University of New York, New Paltz, New York 12561, USA}
		\city{Xi'an}
		\country{China}}
	\email{lik@newpaltz.edu}

	\renewcommand{\shortauthors}{Shou et al.}
	
	\begin{abstract}
	Multi-modal Emotion Recognition in Conversation (MERC) has received considerable attention in various fields, e.g., human-computer interaction and recommendation systems. Most existing works perform feature disentanglement and fusion to extract emotional contextual information from multi-modal features and emotion classification. After revisiting the characteristic of MERC, we argue that long-range contextual semantic information should be extracted in the feature disentanglement stage and the inter-modal semantic information consistency should be maximized in the feature fusion stage. Inspired by recent State Space Models (SSMs), Mamba can efficiently model long-distance dependencies. Therefore, in this work, we fully consider the above insights to further improve the performance of MERC. Specifically, on the one hand, in the feature disentanglement stage, we propose a Broad Mamba, which does not rely on a self-attention mechanism for sequence modeling, but uses state space models to compress emotional representation, and utilizes broad learning systems to explore the potential data distribution in broad space. Different from previous SSMs, we design a bidirectional SSM convolution to extract global context information. On the other hand, we design a multi-modal fusion strategy based on probability guidance to maximize the consistency of information between modalities. Experimental results show that the proposed method can overcome the computational and memory limitations of Transformer when modeling long-distance contexts, and has great potential to become a next-generation general architecture in MERC. 
	\end{abstract}

	\ccsdesc[500]{Computing methodologies~Discourse, dialogue and pragmatics}
	\ccsdesc[300]{Computing methodologies~Non-negative matrix factorization}
	\ccsdesc[100]{Theory of computation~Fixed parameter tractability}
	
	\keywords{Multi-modal Emotion Recognition, State Space Model, Broad Learning System, Feature Fusion, Probabilistic Guidance}
	
	\maketitle
	
\section{Introduction}

\begin{figure}
	\centering
	\includegraphics[width=1\linewidth]{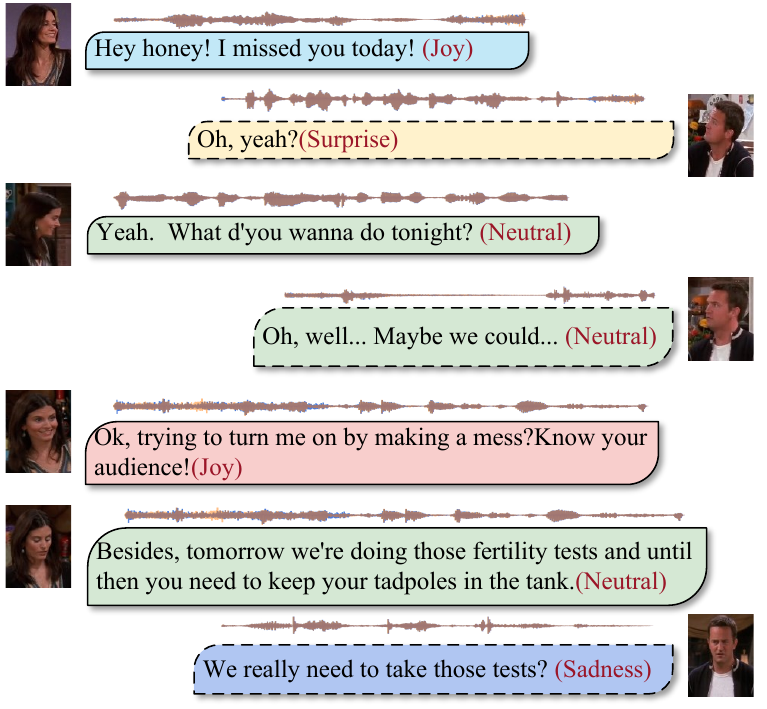}
	\caption{An illustrative example of multi-modal emotion recognition in conversation. For each given sentence, it contains three modal information about the speaker, i.e., text, video and audio. The task of MERC is to identify the emotional labels contained in the three modal information.}
	\label{fig:shouming}
\end{figure}

Emotion recognition in conversation \cite{xing2020adapted, gandhi2023multimodal, ai2023two, shou2023adversarial, meng2024multi} has received considerable research attention and has been widely used in various fields, e.g., emotion analysis \cite{hu2021mmgcn} and public opinion warning \cite{yan2022research}, etc. Recently, research on Multi-modal Emotion Recognition in Conversation (MERC) has mainly focused on multimodality, i.e., text, video and audio \cite{lu2020iterative, ren2021interactive, chen2021learning, deshmukh2023pengi, meng2023deep, shou2023comprehensive}. As shown in Fig. \ref{fig:shouming}, MERC aims to identify emotion labels in sentences with text, video, and audio information. Unlike previous work \cite{kim-2014-convolutional} that only uses text information for emotion recognition, MERC improves the model's emotion understanding capabilities by introducing audio and video information \cite{hou2023semantic, shou2022conversational}. The introduction of audio and video alleviates the limitation of insufficient semantic information caused by relying solely on text features. 

Many existing works \cite{liu2023multi, sun2023layer, zhang2023structure} improve the performance of MERC by effectively extracting contextual semantic information of different modalities and fusing inter-modal complementary semantic information. By revisiting the characteristics of MERC, we argue that the core idea of MERC includes a feature disentanglement step and a feature fusion step.

Specifically, the goal of feature disentanglement is to extract the contextual semantic information most relevant to emotional features in multi-modal features \cite{wang2024inter, yu2020ch}. Recent work on Transformers \cite{fan2023mgat, lian2021ctnet, li2023skier} has achieved great success in modeling long-range contextual semantic information. Compared with traditional Recurrent Neural Networks (RNNs) \cite{majumder2019dialoguernn, li2022bieru}, the advantage of Transformer is that it can effectively provide global contextual semantic information through the attention mechanism in parallel. However, the quadratic complexity of the self-attention mechanism in Transformers poses challenges in terms of speed and memory when dealing with long-range context dependencies. Inspired by the state space models, Mamba with linear complexity is proposed to achieve efficient training and inference. Mamba's excellent scaling performance shows that it is a promising Transformer alternative for context modeling. Therefore, to efficiently extract long-distance contextual semantic information, we designed the broad Mamba, which incorporates the SSMs for data-dependent global emotional context modeling, and a broad learning system to explore the potential data distribution in the broad space. Different from previous SSMs, we design a bidirectional SSM convolution to extract global context information. In addition, we also introduce position encoding information to improve SSMs' ability to understand sequences at different positions.

After completing feature disentanglement, the model needs to perform feature fusion to maximize the consistency of information between different modalities. The core idea of feature fusion is to assign different weights by determining the importance of different modal features to downstream tasks. Many cross-modal feature fusions have been proposed in existing MERC research, e.g., tensor fusion network \cite{zadeh2017tensor}, graph fusion network \cite{zhang2023dualgats}, attention fusion \cite{ren2021interactive}. However, the feature fusion process in previous works is relatively coarse-grained and cannot actually determine the contribution of each modal feature to downstream tasks. We argue that label information plays an important role in guiding multi-modal information fusion. Therefore, how to properly fuse multi-modality and determine the contribution of multi-modal features to downstream tasks in a fine-grained manner remains a challenge \cite{ju2024survey, meng2024revisiting}.

To tackle the above problems, we propose an effective probability-guided fusion mechanism to achieve multi-modal contextual feature fusion, which utilizes the predicted label probability of each modal feature as the weight vectors of the modal features. Compared with other feature fusion models for emotion recognition tasks, the proposed fusion method can utilize the predicted label probability information in a fine-grained manner to actually determine the contribution of different modal features to the emotion prediction task.

To evaluate the effectiveness and efficiency of our proposed method, we conduct extensive experiments on two widely used benchmark datasets, IEMOCAP and MELD. In fact, the proposed method achieves state-of-the-art performance with low computational consumption, and experimental results demonstrate its effectiveness and efficiency.

Overall, our main contributions can be summarized as follows:

\begin{itemize}
	\item We propose a Broad Mamba, which combines a broad learning system for searching abstract emotional features in a broad space and a SSM for data-dependent global emotional context information extraction. Different from previous SSMs, we design a bidirectional SSM convolution to extract global context information.
	
	\item We propose an effective probability-guided fusion mechanism to achieve multi-modal contextual feature fusion, which utilizes the predicted label probability of each modal feature as the weight vectors of the modal features.
	
	\item We conduct extensive experiments on the IEMOCAP and MELD datasets. Experimental results show that our proposed method achieves superior performance compared with the well-established Transformer or GNN architectures.
\end{itemize}
	
\section{Related work}

\subsection{Multi-modal Emotion Recognition in Conversation}
In the early eras, GRU \cite{chung2014empirical} and LSTM \cite{6795963} are the de-facto standard network designs for Natural Language Processing (NLP). Many recurrent neural network architectures \cite{hazarika2018icon, majumder2019dialoguernn, jiao2020real, li2022bieru, hazarika2018conversational} have been proposed for various Multi-modal Emotion Recognition in Conversation (MERC). The pioneering work, Transformer changed the landscape by enabling efficient parallel computing under the premise of long sequence modeling. Transformer treats text as a series of 1D sequence data and applies an attention architecture to achieve sequence modeling. Transformer's surprising results on long sequence modeling and its scalability have encouraged considerable follow-up work for MERC \cite{chudasama2022m2fnet, lian2021ctnet, shen2021dialogxl, zhong2019knowledge, li2022emocaps}. One line of works focus on achieving intra-modal and inter-modal information fusion. For example, CTNet \cite{lian2021ctnet} proposes a single Transformer and cross Transformer. CKETF \cite{ghosh2021context} constructs a Context and Knowledge Enriched Transformer. TL-ERC applies the Transformer with the transfer learning. Another pioneering work, Graph Neural Network (GNN) \cite{yin2023coco, yin2023messages, yin2023omg} further improved the performance of ERC. The core idea of GNN is to learn the representation of nodes or graphs through the feature information of nodes and the connection relationships in the graph structure \cite{chiang2019cluster, min2020scattering, zhang2023dualgats}. For instance, DialogueGCN \cite{ghosal2019dialoguegcn} proposes to use context information to build dialogue graphs. DER-GCN \cite{ai2024gcn} fuses event relationships into speaker relationship graphs.

These dominant follow-up works have demonstrated excellent performance and higher efficiency on various multi-modal conversational emotion recognition data sets by introducing attention mechanisms or GNNs. In this work, we draw inspiration from Mamba and explore the ability to build a state space model (SSM) based model to improve multi-modal emotion representation learning without efficient parallel sequence modeling using attention, while retaining the sequence modeling advantages of Transformer.

\subsection{State Space Models}
The State Space Models (SSMS) is used to describe the dynamic change process consisting of observed values and unknown internal state variables.	Gu et al. \cite{gu2021efficiently} proposes a Structured State Space Sequence (S4) model, an alternative to the Transformer architecture that models long-range dependencies without using attention. The property of linear complexity of state space sequence lengths has received considerable research attention. Smith et al. \cite{smith2022simplified} improves S4 by introducing MIMO SSM and efficient parallel scanning into the S4 layer to achieve parallel initialization and state reset of the hidden layer. He et al. \cite{he2024densemamba} proposes introducing dense connection layers into SSM to improve the feature representation ability of shallow hidden layer states. Mehta et al. \cite{mehta2023long} improves the memory ability of the hidden layer by introducing gated units on S4. Recently, Gu et al. \cite{gu2023mamba} proposes the general language model Mamba, which has better sequence modeling capabilities than Transformers and is linearly complex. Zhu et al. \cite{zhu2024vision} introduces bidirectional SSM based on Mamba to improve the context information representation of the hidden layer. In this work, we are inspired by Mamba to transfer SSM to emotion representation learning without attention computation.
	
\section{Preliminary Information}
	
\subsection{Multi-modal Feature Extraction}
	
\textbf{Word Embedding:} Following previous work \cite{ma2023transformer, li2023revisiting}, we use RoBerta in this paper to obtain context-embedded representations of text. Specifically, we first segment the input text and add the start symbol `[CLS]' and the end symbol `[SEP]'. The processed input data is then passed to the RoBERTa model to obtain contextual representations $\boldsymbol{\xi}_t$ of the text.
	
\textbf{Visual and Audio Feature Extraction:} For video and audio features, following previous work \cite{li2023revisiting, fan2023mgat}, we utilize DenseNet and openSMILE for feature extraction and obtain video embedding features $\boldsymbol{\xi}_v$ and audio embedding features  $\boldsymbol{\xi}_a$, respectively.

\subsection{State Space Model}
The State Space Model (SSMs) is an efficient sequence modeling model that can capture the dynamic changes of data over time. Owing to the efficient sequence modeling capabilities, SSM has received widespread attention in various fields, e.g., video understanding and image segmentation. A typical SSM consists of a state equation and an observation equation, where the state equation describes the dynamic changes within the system, and the observation equation describes the connection between the system state and observations. Given an input $x(t) \in \mathbb{R}$ and a hidden state $h(t) \in \mathbb{R}$, $y(t)$ is obtained mathematically through a linear ordinary differential equations (ODE) as follows:
\begin{equation}
	\label{eq:1}
	\begin{aligned}
		h'(t)&=\mathbf{A}h(t)+\mathbf{B}x(t),\\y(t)&=\mathbf{C}h(t)
	\end{aligned}
\end{equation}
where $\mathbf{A} \in \mathbb{R}^{N \times N}$ is the evolution parameter and $\mathbf{B} \in \mathbb{R}^{N \times 1}, \mathbf{C} \in \mathbb{R}^{1 \times N}$ are the projection parameters, and $N$ is the latent state size.

SSM is a continuous time series model, which is difficult to efficiently integrate into deep learning algorithms. Inspired by SSM, Mamba discretizes ODEs to achieve computational efficiency. Mamba discretizes the evolution parameter $\mathbf{A}$ and the projection parameter $\mathbf{B}$ by introducing a timescale parameter $\boldsymbol{\Delta}$ to obtain $\overline{\mathbf{A}}$ and $\overline{\mathbf{B}}$. The formula is defined as follows:
\begin{equation}
	\begin{aligned}&\overline{\mathbf{A}}=\exp{(\boldsymbol{\Delta}\mathbf{A})},\\&\overline{\mathbf{B}}=(\boldsymbol{\Delta}\mathbf{A})^{-1}(\exp{(\boldsymbol{\Delta}\mathbf{A})}-\mathbf{I})\cdot\boldsymbol{\Delta}\mathbf{B}
\end{aligned}\end{equation}

In practice, we use a first-order Taylor series to obtain an approximation of $\overline{\mathbf{B}}$ as follows:
\begin{equation}\overline{\mathbf{B}}=(e^{\boldsymbol{\Delta} \mathbf{A}}-\mathbf{I})\mathbf{A}^{-1}\mathbf{B}\approx(\boldsymbol{\Delta} \mathbf{A})(\boldsymbol{\Delta} \mathbf{A})^{-1}\boldsymbol{\Delta} \mathbf{B}=\boldsymbol{\Delta} \mathbf{B}\end{equation}

After obtaining the discretized $\overline{\mathbf{A}}$ and $\overline{\mathbf{B}}$, we rewrite Eq. \ref{eq:1} as follows:
\begin{equation}
	\begin{aligned}
		h_t&=\overline{\mathbf{A}}h_{t-1}+\overline{\mathbf{B}}x_t,\\y_t&=\mathbf{C}h_t + \mathbf{D}x_t
	\end{aligned}
\end{equation}
and then the output is computed via global convolutiona as follows:
\begin{equation}
	\label{eq:5}
	\begin{aligned}\overline{\mathbf{K}}&=(\mathbf{C}\overline{\mathbf{B}},\mathbf{C}\overline{\mathbf{A}\mathbf{B}},\ldots,\mathbf{C}\overline{\mathbf{A}}^{\text{M}}\overline{\mathbf{B}}),\\\mathbf{y}&=\mathbf{x}*\overline{\mathbf{K}} + \mathbf{x}*\mathbf{D}
\end{aligned}\end{equation}

We adopted Mamba as a sequence modeling method in this work since Mamba can efficiently process sequence data without significant performance degradation.

\subsection{Broad Learning System}
\begin{figure}
	\centering
	\includegraphics[width=1\linewidth]{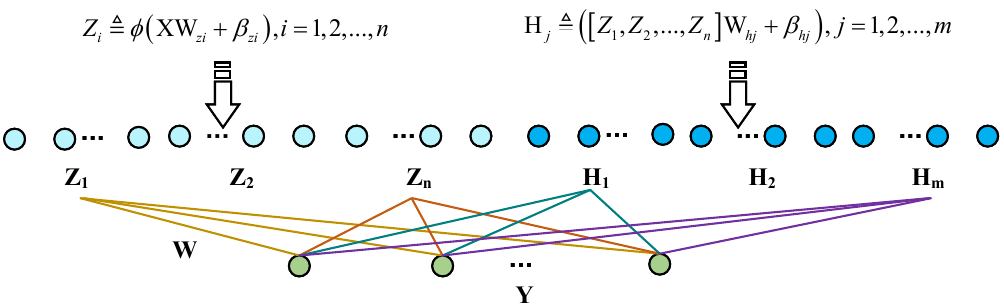}
	\caption{The overall architecture of Broad Learning System (BLS). $\mathbf{Z}_i$ represents the feature nodes, $\mathbf{H}_i$ represents the enhancement nodes, and $\mathbf{Y}$ represents the predicted labels.}
	\label{fig:broad-learning-systems}
\end{figure}

Broad Learning System (BLS) is different from traditional deep learning methods that it mainly focuses on discovering the relationship between features in the input data, rather than extracting features through multi-level nonlinear transformations. The core idea of BLS is to jointly solve the optimization problem by integrating the semantic information of feature nodes and enhancement nodes. Notably, the feature nodes and enhancement nodes only contain one layer of learnable network parameters, so BLS has faster inference speed than other deep learning architectures. The overall process of the BLS algorithm is shown in the Fig. \ref{fig:broad-learning-systems}.

\begin{figure*}
	\centering
	\includegraphics[width=1\linewidth]{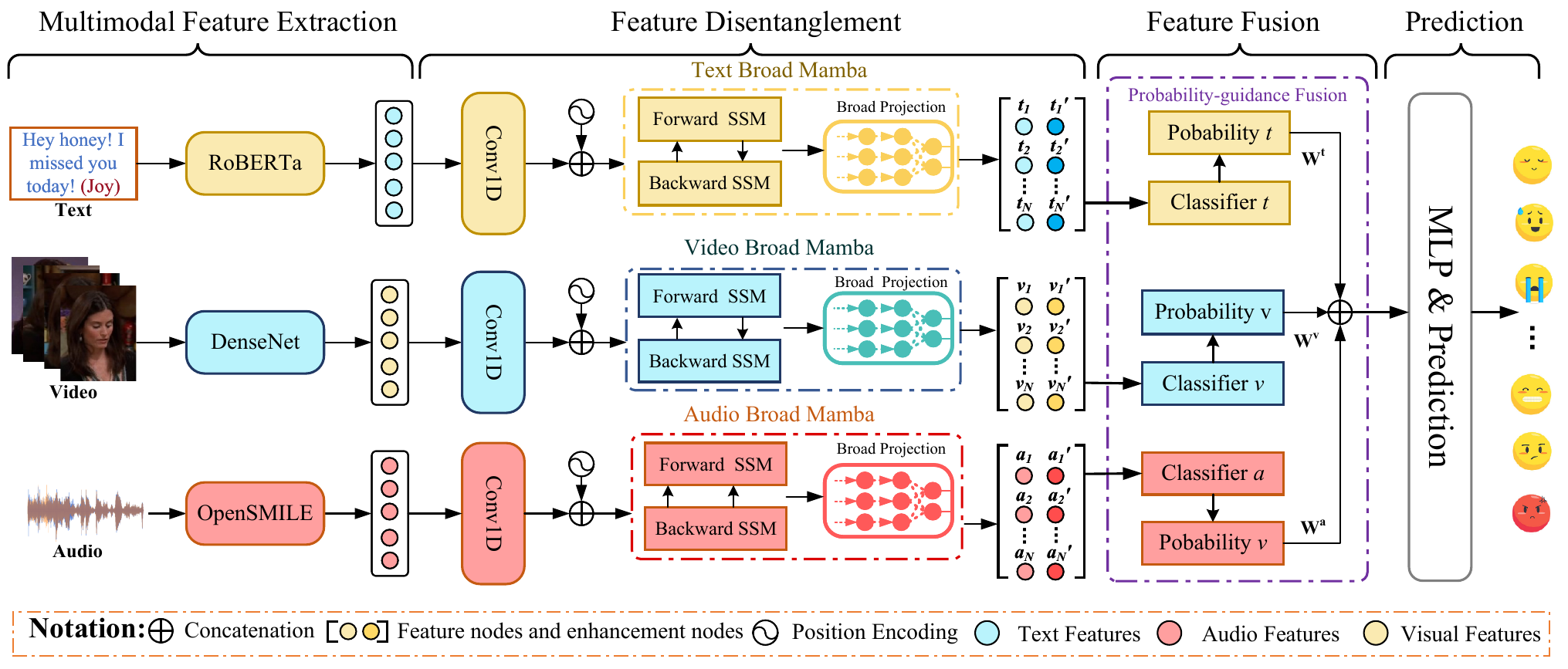}
	\caption{The overall framework of the proposed model. Specifically, we first input the extracted multi-modal features into a 1-D convolutional layer for multi-scale feature extraction and introduce position encoding information to consider the position information of the series in the context. Then we input the obtained multi-modal features with multi-scale information into Broad Mamba to extract contextual semantic information and explore the potential data distribution in the broad space. Finally, we use a probability-guidance fusion model to complete the fusion of multi-modal features and achieve emotion prediction.}
	\label{fig:mamba}
\end{figure*}

Specifically, for a given input data $\mathbf{X} \in \mathbb{R}^{N \times M}$, where $N$ represents the number of samples and $M$ represents the dimension of the feature. The generated feature nodes are defined as follows:
\begin{equation}
	\mathbf{Z}_i\triangleq\phi(\mathbf{X}\mathbf{W}_{z_i}+\boldsymbol{\beta}_{z_i}), i= 1, 2, \ldots, n
\end{equation}
where $\mathbf{W}_{z_i} \in \mathbb{R}^{M \times d_{z}}$ and $\beta_{z} \in \mathbb{R}^{1 \times d_z}$ are the learnable parameters. $d_z$ is the embedding dimensions of generated features and $\phi$ is the activation function. The set of generated feature nodes is represented as $\mathbf{Z}^n\triangleq[\mathbf{Z}_1,\ldots,\mathbf{Z}_n]$, where $n$ is the size of the set of generated feature nodes. Similarly, enhancement node features are defined as follows:

\begin{equation}
	\mathbf{H}_j\triangleq\phi(\mathbf{Z}\mathbf{W}_{h_j}+\boldsymbol{\beta}_{h_j}), j=1,2,\ldots,m
\end{equation}
where $\mathbf{W}_{h_i} \in \mathbb{R}^{d_{z} \times d_h}$ and $\beta_{z} \in \mathbb{R}^{1 \times d_h}$ are the learnable parameters. $d_h$ is the embedding dimensions of enhancement features. The set of enhancement feature nodes is represented as $\mathbf{H}^m\triangleq[\mathbf{H}_1,\ldots,\mathbf{H}_m]$, where $m$ is the size of the set of enhancement feature nodes.

The final model output by concatenating feature nodes and enhancement nodes is as follows:
\begin{equation}
	\begin{aligned}\mathbf{Y}&=[\mathbf{Z}_1,\ldots,\mathbf{Z}_n|\mathbf{H}_1,\ldots,\mathbf{H}_m]\mathbf{W}\\&=[\mathbf{Z}^n|\mathbf{H}^m]\mathbf{W}
\end{aligned}
\end{equation}
where $\mathbf{W}$ is the learnable parameters.

\section{The proposed method}

\subsection{Feature Disentanglement}

\subsubsection{1D-Conv} To capture features of different scales and abstraction levels in multi-modal features (e.g., information such as the relationship between words and the importance of utterence), we input text features $\boldsymbol{\xi}_t$, video features $\boldsymbol{\xi}_v$ and audio features $\boldsymbol{\xi}_a$ into a 1D convolutional network (Conv1D) as follows:

\begin{equation}
	\hat{\boldsymbol{\xi}}_t/\hat{\boldsymbol{\xi}}_a/\hat{\boldsymbol{\xi}}_v=Conv1D_{t/a/v}\left(\boldsymbol{\xi}_t,\boldsymbol{\xi}_a,\boldsymbol{\xi}_v\right)
\end{equation}
where $\hat{\boldsymbol{\xi}}_t \in \mathbb{R}^{T_t \times d_m}$, $\hat{\boldsymbol{\xi}}_a \in \mathbb{R}^{T_a \times d_m}$, and $\hat{\boldsymbol{\xi}}_v \in \mathbb{R}^{T_v \times d_m}$, $T_t, T_a, T_v$ represent the feature dimensions of text, audio, and video respectively, $d_m$ represents the output feature dimensions.

Furthermore, to facilitate the model to capture the dependencies between long-distance positions in the sequence, we introduce sine and cosine position encoding embedding as follows:
\begin{equation}
	\begin{aligned}
		PE_{(pos,2i)}&=\sin\left(\frac{pos}{10000^{2i/d}}\right)\\PE_{(pos,2i+1)}&=\cos\left(\frac{pos}{10000^{2i/d}}\right)\end{aligned}
\end{equation}
where $pos$ represents the position in the sequence. $i$ represents the dimension index of position encoding, $i = 0, 1, ..., D-1$. $D$ represents the embedded dimension. We input $\boldsymbol{\hat{\xi}}_t,\boldsymbol{\hat{\xi}}_a,and \boldsymbol{\hat{\xi}}_v$ ($\boldsymbol{\hat{\xi}}_t,\boldsymbol{\hat{\xi}}_a, \boldsymbol{\hat{\xi}}_v = Conv1D_{t/a/v}\left(\boldsymbol{\xi}_t,\boldsymbol{\xi}_a,\boldsymbol{\xi}_v\right) + PE$) that encodes position information at each time step into Broad Mamba.

\begin{figure}
	\centering
	\includegraphics[width=1\linewidth]{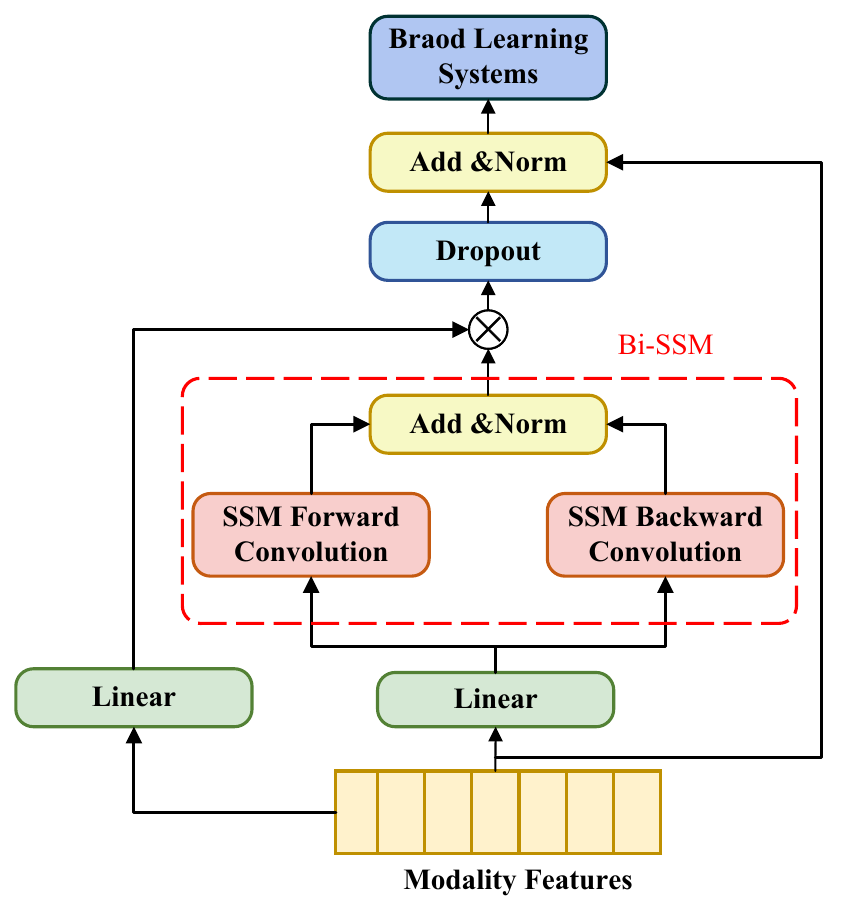}
	\caption{The overall architecture of Broad Mamba. We use a bidirectional SSM to encode forward and reverse contextual semantic information.}
	\label{fig:broad-mamba}
\end{figure}

\subsubsection{Broad Mamba}
The overall architecture of the proposed Broad Mamba  is shown in Fig. \ref{fig:broad-mamba}. In order to aggregate the contextual semantic information from the forward and backward directions, we build a bidirectional SSM convolution module. Specifically, the first kernel $\overleftarrow{\boldsymbol{\kappa}}$ performs a 1D convolution operator to obtain forward context information. The second kernel $\overrightarrow{\boldsymbol{\kappa}}$ performs a 1D convolution operator to obtain the mutual information associated with emotional information, and we add the two convolved results. The overall operating process is formally defined as follows:

\begin{equation}
	\begin{aligned}
		\boldsymbol{\bar{\xi}}_j^{t/a/v}&=\sum_{l\leq j}\overleftarrow{\boldsymbol{\kappa}}_{j-l}^{t/a/v}\odot\boldsymbol{\hat{\xi}}_l^{t/a/v}+\sum_{l\geq j}\overrightarrow{\boldsymbol{\kappa}}_{l-j}^{t/a/v}\odot\boldsymbol{\hat{\xi}}_l^{t/a/v}+\boldsymbol{d}^{t/a/v}\odot\boldsymbol{\hat{\xi}}_j^{t/a/v}\\&=\text{BiSSM}(\boldsymbol{\hat{\xi}}_j^{t/a/v})
	\end{aligned}
\end{equation}
where $\overleftarrow{\boldsymbol{\kappa}}$, and $\overrightarrow{\boldsymbol{\kappa}}$ are obtained via Eq. \ref{eq:5}.

To explore the potential data distribution of multi-modal data in the broad space and improve the performance of Mamba, we use Broad Learning Sytems (BLS) to enhance the emotional representation ability of features. Specifically, we map the features output by BiSSM to a random broad space and obtain feature nodes and enhancement nodes, and concatenate the feature nodes and enhancement nodes as the input of the feature fusion layer. Specifically, feature nodes can be formally defined as follows:

\begin{equation}\mathbf{Z}_i^{t/a/v}\triangleq \text{BiSSM}(\boldsymbol{\hat{\xi}}_j^{t/a/v})\mathbf{W}_{z_i}^{t/a/v}+\boldsymbol{\beta}_{z_i}^{t/a/v},\mathrm{~}i=1,2,\ldots,n,\end{equation}
and the enhancement nodes can be computed as:
\begin{equation}
	\mathbf{H}_j^{t/a/v}\triangleq\mathrm{ReLU}(\mathbf{Z}^n_{t/a/v}\mathbf{W}_{h_j}^{t/a/v}+\boldsymbol{\beta}_{h_j}^{t/a/v}), j=1,2,\ldots,m.
\end{equation}

Furthermore, we introduce l2 regularization into the loss function to avoid the overfitting phenomenon of BLS, which is formally defined as follows:
\begin{equation}
	\label{eq:14}
	\mathcal{L}_{norm}=\|[\mathbf{Z}^n_{t/a/v}|\mathbf{H}^m_{t/a/v}]\mathbf{W}_b^{t/a/v}-\mathbf{Y}^{t/a/v}\|_2^2+\lambda\|\mathbf{W}_b^{t/a/v}\|_2^2
\end{equation}
where $\lambda$ is the weight decay coefficient, $\mathbf{W}_b^{t/a/v}$ is the learnable parameters, $\mathbf{Y}^{t/a/v}=[\mathbf{Z}_1^{t/a/v},\ldots, \mathbf{Z}_n^{t/a/v}, \ldots, \mathbf{H}_1^{t/a/v}, \ldots, \mathbf{H}_n^{t/a/v}]$.

By deriving and solving the Eq. \ref{eq:14}, the solution to $\mathbf{W}_b^{t/a/v}$ can be calculated as follows:
\begin{equation}
	\begin{aligned}
	\mathbf{W}_b^{t/a/v}=&\left([\mathbf{Z}^n_{t/a/v}|\mathbf{H}^m_{t/a/v}]^\top[\mathbf{Z}^n_{t/a/v}|\mathbf{H}^m_{t/a/v}]+\lambda\mathbf{I}\right)^{-1} \\ &[\mathbf{Z}^n_{t/a/v}|\mathbf{H}^m_{t/a/v}]^\top\mathbf{Y}^{t/a/v}
	\end{aligned}
\end{equation}

\begin{table*}[htbp]
	\setlength{\tabcolsep}{6.5pt}
	\centering
	\caption{Comparison with other baseline models on the IEMOCAP dataset. The best result in each column is in bold.}
	\label{tab:iemocap}
	\begin{tabular}{@{}l|c|cccccccccccccc@{}}
		\toprule
		\multirow{3}{*}{Methods} & \multirow{3}{*}{Parmas.} & \multicolumn{14}{c}{IEMOCAP}                                                                                                                                                                                  \\ \cmidrule(l){3-16} 
		&                          & \multicolumn{2}{c}{Happy} & \multicolumn{2}{c}{Sad} & \multicolumn{2}{c}{Neutral} & \multicolumn{2}{c}{Angry} & \multicolumn{2}{c}{Excited} & \multicolumn{2}{c}{Frustrated} & \multicolumn{2}{c}{Average(w)} \\
		&                          & Acc.        & F1          & Acc.       & F1         & Acc.         & F1           & Acc.        & F1          & Acc.         & F1           & Acc.           & F1            & Acc.           & F1            \\ \midrule
		bc-LSTM                  &    1.28M                      & 29.1        & 34.4        & 57.1       & 60.8       & 54.1         & 51.8         & 57.0        & 56.7        & 51.1         & 57.9         & 67.1           & 58.9          & 55.2           & 54.9          \\
		LFM                      &   2.34M                       & 25.6        & 33.1        & 75.1       & 78.8       & 58.5         & 59.2         & 64.7        & 65.2        & 80.2         & 71.8         & 61.1           & 58.9          & 63.4           & 62.7          \\
		A-DMN                    &       --                   & 43.1        & 50.6        & 69.4       & 76.8       & 63.0         & 62.9         & 63.5        & 56.5        & \textbf{88.3}         & 77.9         & 53.3           & 55.7          & 64.6           & 64.3          \\
		DialogueGCN              &     12.92M                     & 40.6        & 42.7        & \textbf{89.1}       & \textbf{84.5}       & 62.0         & 63.5         & 67.5        & 64.1        & 65.5         & 63.1         & 64.1           & 66.9          & 65.2           & 64.1          \\
		RGAT                     &    15.28M                      & 60.1        & 51.6        & 78.8       & 77.3       & 60.1         & 65.4         & 70.7        & 63.0        & 78.0         & 68.0         & 64.3           & 61.2          & 65.0           & 65.2          \\
		CoMPM                    &       --                   & 59.9        & 60.7        & 78.0       & 82.2       & 60.4         & 63.0         & 70.2        & 59.9        & 85.8         & 78.2         & 62.9           & 59.5          & 67.7           & 67.2          \\
		EmoBERTa                 & 499M                         & 56.9        & 56.4        & 79.1       & 83.0       & 64.0         & 61.5         & 70.6        & 69.6        & 86.0         & 78.0         & 63.8           & 68.7          & 67.3           & 67.3          \\
		COGMEN                   &   --                       & 57.4        & 51.9        & 81.4       & 81.7       & 65.4         & 68.6         & 69.5        & 66.0        & 83.3         & 75.3         & 63.8           & 68.2          & 68.2           & 67.6          \\
		CTNet                    &    8.49M                      & 47.9        & 51.3        & 78.0       & 79.9       & 69.0         & 65.8         & \textbf{72.9}        & 67.2        & 85.3         & \textbf{78.7}         & 52.2           & 58.8          & 68.0           & 67.5          \\
		LR-GCN                   &    15.77M                      & 54.2        & 55.5        & 81.6       & 79.1       & 59.1         & 63.8         & 69.4        & 69.0        & 76.3         & 74.0         & 68.2           & 68.9          & 68.5           & 68.3          \\
		AdaGIN       & 6.3M  & 53.0        &   --    & 81.5    &  --  &  71.3   &  -- & 65.9   & -- &  76.3  &  -- & 67.8 &  -- & 70.5  & 70.7  \\
		DER-GCN                  &  78.59M                        & \textbf{60.7}        & 58.8        & 75.9       & 79.8       & 66.5         & 61.5         & 71.3        & \textbf{72.1}        & 71.1         & 73.3         & 66.1           & 67.8          & 69.7           & 69.4         \\  \hdashline
		Our Model & 1.73M & 58.1 & \textbf{65.5} & 86.0 & 81.6 & \textbf{70.7} & \textbf{73.5} & 69.1 & 70.1 & 85.5 & 76.3 & \textbf{69.5} & \textbf{69.8} & \textbf{73.1} & \textbf{73.3} \\ \bottomrule
	\end{tabular}
\end{table*}

\subsubsection{Computation-Efficiency} SSM and the self-attention mechanism in Transformer both plays an important role in modeling global contextual semantic information. However, the self-attention mechanism is quadratic in complexity and is very time-consuming in training and inference. On the contrary, the computational complexity of SSM is $O(Llog L)$, so it can accelerate model inference in modeling long sequences.

\subsection{Feature Fusion}

\subsubsection{Probability-guided Fusion Model}
Many studies have proven that different modalities have different contributions to the prediction of emotional labels, so modal features with higher contributions need to be given greater weight in the multi-modal feature fusion process. Different from previous works that fuse modal features at a coarse-grained level without using label information for guidance, we design a probability-guided fusion model (PFM) that dynamically assigns weights to each modality by using the predicted emotion label probabilities of the modalities. Specifically, we build an emotion classifier for the feature representation of each modality to obtain the predicted probability of the label as the weight of the modal features in the fusion process. The fusion process is formally defined as follows:
\begin{equation}
	\omega^{t/a/v}=\mathrm{Sigmoid}\left(\mathrm{MLP}^{t/a/v}\left(\mathrm{Y}^{t/a/v}\right)\right)
\end{equation}
and then we can obtain the fused multi-modal feature representations as follows:
\begin{equation}h^f=\omega^t\mathrm{Y}^t+\omega^a\mathrm{Y}^a+\omega^v\mathrm{Y}^v\end{equation}

\begin{table*}[htbp]
	\caption{Comparison with other baseline models on the MELD dataset. The best result in each column is in bold.}
	\label{tab:meld}
	\setlength{\tabcolsep}{4.9pt}{
		\begin{tabular}{@{}l|c|cccccccccccccccl@{}}
			\toprule
			\multirow{3}{*}{Methods} & \multirow{3}{*}{Parmas.} & \multicolumn{16}{c}{MELD}                                                                                                                                                                                                             \\ \cmidrule(l){3-18} 
			&                          & \multicolumn{2}{c}{Neutral} & \multicolumn{2}{c}{Surprise} & \multicolumn{2}{c}{Fear} & \multicolumn{2}{c}{Sadness} & \multicolumn{2}{c}{Joy} & \multicolumn{2}{c}{Disgust} & \multicolumn{2}{c}{Anger} & \multicolumn{2}{c}{Average(w)} \\
			&                          & Acc.         & F1           & Acc.          & F1           & Acc.        & F1         & Acc.         & F1           & Acc.       & F1         & Acc.         & F1           & Acc.        & F1          & Acc.  & \multicolumn{1}{c}{F1} \\ \midrule
			bc-LSTM                  &    1.28M                      & 78.4         & 73.8         & 46.8          & 47.7         & 3.8         & 5.4        & 22.4         & 25.1         & 51.6       & 51.3       & 4.3          & 5.2          & 36.7        & 38.4        & 57.5  & 55.9                   \\
			DialogueRNN              &  14.47M                        & 72.1         & 73.5         & 54.4          & 49.4         & 1.6         & 1.2        & 23.9         & 23.8         & 52.0       & 50.7       & 1.5          & 1.7          & 41.0        & 41.5        & 56.1  & 55.9                   \\
			DialogueGCN              &  12.92M                        & 70.3         & 72.1         & 42.4          & 41.7         & 3.0         & 2.8        & 20.9         & 21.8         & 44.7       & 44.2       & 6.5          & 6.7          & 39.0        & 36.5        & 54.9  & 54.7                   \\
			RGAT                     &  15.28M                        & 76.0         & 78.1         & 40.1          & 41.5         & 3.0         & 2.4        & 32.1         & 30.7         & 68.1       & 58.6       & 4.5          & 2.2          & 40.0        & 44.6        & 60.3  & 61.1                   \\
			CoMPM                    &   --                       & 78.3         & 82.0         & 48.3          & 49.2         & 1.7         & 2.9        & 35.9         & 32.3         & 71.4       & 61.5       & 3.1          & 2.8          & 42.2        & 45.8        & 64.1  & 65.3                   \\
			EmoBERTa                 &   499M                       & \textbf{78.9}         & \textbf{82.5}         & 50.2          & 50.2         & 1.8         & 1.9        & 33.3         & 31.2         & 72.1       & 61.7       & 9.1          & 2.5          & 43.3        & 46.4        & 64.1  & 65.2                   \\
			A-DMN                    &      --                    & 76.5         & 78.9         & 56.2          & 55.3         & 8.2         & 8.6        & 22.1         & 24.9         & 59.8       & 57.4       & 1.2          & 3.4          & 41.3        & 40.9        & 61.5  & 60.4                   \\
			LR-GCN                   &     15.77M                     & 76.7         & 80.0         & 53.3          & 55.2         & 0.0         & 0.0        & 49.6         & 35.1         & 68.0       & \textbf{64.4}       & 10.7         & 2.7          & 48.0        & 51.0        & 65.7  & 65.6                   \\
			AdaGIN       & 6.3M  & 79.8        &   --    & 60.5    & --   &  15.2   & --  & 43.7   & -- &  64.5  & --  & 29.3 & -- & 56.2 &  -- & 67.6  & 66.8  \\
			DER-GCN                  &   78.59M                       & 76.8         & 80.6         & 50.5          & 51.0         & 14.8        & \textbf{10.4}       & \textbf{56.7}         & 41.5         & 69.3       & 64.3       & 17.2         & 10.3         & 52.5        & \textbf{57.4}        & 66.8  & 66.1     \\ \hdashline              
			Our model & 1.73M  & 73.1 & 79.7 & \textbf{62.5}  & \textbf{60.6}   & \textbf{25.0}  & 6.9   & 52.9   & 38.9   & 76.1  & 63.0  & \textbf{57.1}    & \textbf{27.0}  & \textbf{53.0}  & 57.1 & \textbf{68.0} &  \textbf{67.6} \\ \bottomrule
		\end{tabular}}
	\end{table*}

\subsection{Model Training}

Finally, we utilize the fused multi-modal emotional context features for emotion classification. The formula is defined as follows:
\begin{equation}
	\hat{y}_i=\mathrm{argmax}\left(\mathrm{softmax}\left(\mathrm{MLP}(h^f)\right)\right)
\end{equation}
where $\hat{y}_i$ is the predicted emotion labels, $\mathrm{MLP}$ represents the multilayer perceptron with multiple layers of learnable parameters. We use cross-entropy loss as the classification loss of the model, and the formula is defined as follows:

\begin{equation}
	\mathcal{L}_{emo}=-\sum_{i=1}^ny_i\log{\hat{y}}_{i}
\end{equation}
where $y_i$ is the true emotion labels, $n$ is the number of the samples.

Therefore, during the optimization phase of the model, the overall training loss function is defined as follows:
\begin{equation}
	\mathcal{L} = \mathcal{L}_{norm} + \mathcal{L}_{emo}
\end{equation}

\section{EXPERIMENTS}
In this section, we firstly introduce the experimental data set and evaluation metrics. Then the baselines method in the experiments are explained. We then compare the proposed method with baseline work on two benchmark datasets. Secondly, we conduct ablation studies to analyze the effectiveness of the proposed module and the importance of multi-modal features. Finally, we use tsne to visualize the distribution of the data set. In comparative experiments, our experimental results are the average of 10 runs with different weight initializations. The results of our experiments are statistically significant (all $p < 0.05$) under paired $t$-tests.

\subsection{Implementation Details}
In the experiments, we use AdamW as the optimizer to update the parameters of the network. The model learning rate is set to 1e-4. For the experiment, the number of feature nodes $n$ and the number of enhancement node features $m$ are set to 10 and 30 respectively. Following previous work, we use the same split ratio of training, test, and validation sets for model training and inference.

\begin{table}[htbp]
	\centering
	\caption{Ablation studies for PE, BLS, PFM on the IEMOCAP and MELD datasets.}
	\label{tab:abla}
	\setlength{\tabcolsep}{7pt}{
		\begin{tabular}{@{}l|llll@{}}
			\toprule
			\multirow{2}{*}{Methods} & \multicolumn{2}{c}{IEMOCAP} & \multicolumn{2}{c}{MELD} \\ \cmidrule(l){2-5} 
			& W-Acc.         & W-F1         & W-Acc.        & W-F1       \\ \midrule
			Ours                     &    \textbf{73.1}          &  \textbf{73.3}            &   \textbf{68.0}          &   \textbf{67.6}         \\ \hdashline
			w/o PE                   &  $72.4_{(\downarrow 0.7)}$           &  $72.0_{(\downarrow 1.3)}$         &    $66.7_{(\downarrow 1.3)}$         &    $66.3_{(\downarrow 1.3)}$        \\
			w/o BLS                  &   $71.5_{(\downarrow 1.6)}$          & $  72.1_{(\downarrow 1.2)}$        &  $65.5_{(\downarrow 2.5)}$           &   $64.9_{(\downarrow 2.7)}$         \\
			w/o PFM                  &  $ 70.3_{(\downarrow 2.8)}$          &  $70.7_{(\downarrow 2.6)}$          &    $65.8_{(\downarrow 2.2)}$         & $65.3_{(\downarrow 2.3)}$           \\ \bottomrule
	\end{tabular}}
\end{table}

\begin{table}[htbp]
	\centering
	\setlength{\tabcolsep}{2.5mm}{
		\caption{The effect of our method on IEMOCAP and MELD datasets using unimodal features and multi-modal features, respectively.}
		\label{tab:multi-modal}
		\begin{tabular}{@{}c|llll@{}}
			\toprule
			\multirow{2}{*}{Modality} & \multicolumn{2}{c}{IEMOCAP} & \multicolumn{2}{c}{MELD} \\ \cmidrule{2-5}
			& W-Acc.       & W-F1              & W-Acc         & W-F1         \\ \hline
			T+A+V                     &\textbf{73.1}      & \textbf{73.3}               &    \textbf{68.0}        &  \textbf{67.6} \\ \hdashline
			T                         & $65.5_{(\downarrow 7.6)}$        & $65.7_{(\downarrow 7.6)}$       & $64.6_{(\downarrow 3.4)}$          & $63.9_{(\downarrow 3.7)}$           \\
			A                         & $58.6_{(\downarrow 14.5)}$        & $58.8_{(\downarrow 14.5)}$                & $52.7_{(\downarrow 15.3)}$          & $52.0_{(\downarrow 15.6)}$           \\
			V                       & $49.4_{(\downarrow 23.7)}$        & $49.7_{(\downarrow 23.6)}$                & $40.1_{(\downarrow 27.9)}$          & $41.4_{(\downarrow 26.2)}$
			\\
			T+A                       & $71.3_{(\downarrow 1.8)}$        & $70.2_{(\downarrow 3.1)}$                & $65.2_{(\downarrow 2.8)}$          & $65.6_{(\downarrow 2.0)}$           \\
			T+V                       & $68.7_{(\downarrow 4.4)}$        & $67.4_{(\downarrow 5.9)}$                & $65.0_{(\downarrow 3.0)}$          & $63.7_{(\downarrow 3.9)}$           \\
			V+A                       & $62.1_{(\downarrow 11.0)}$        & $62.2_{(\downarrow 11.1)}$                & 
			$51.3_{(\downarrow 16.7)}$          & $51.9 _{(\downarrow 15.7)}$          \\
			\hline
	\end{tabular}}
\end{table}
	
\subsection{Datasets and Evaluation Metrics}
We conduct experiments using two popular MERC datasets, IEMOCAP \cite{busso2008iemocap} and MELD \cite{poria2019meld}, which include three modal data: text, audio, and video. IEMOCAP contains 12 hours of conversations, each containing six emotion labels. The MELD dataset contains conversation clips from the TV show Friends and contains seven different emotion labels. In addition, in the experiments we report the accuracy (Acc.), and F1 of the proposed method and other baseline methods on each emotion category and the overall weighted average accuracy (W-Acc.), and weighted average F1 (W-F1).

\subsection{Baselines}
We compare several baselines on the IEMOCAP and MELD datasets, including bc-LSTM \cite{poria2017context}, LFM \cite{liu2018efficient}, A-DMN \cite{xing2020adapted}, DialogueGCN \cite{ghosal2019dialoguegcn}, RGAT \cite{ishiwatari2020relation}, CoMPM \cite{lee2022compm}, EmoBERTa \cite{kim2021emoberta}, COGMEN \cite{joshi2022cogmen}, CTNet \cite{lian2021ctnet}, LR-GCN \cite{ren2021lr}, DER-GCN \cite{ai2024gcn}, AdaGIN \cite{tu2024adaptive}.

\subsection{Overall Results}

Tables \ref{tab:iemocap} and \ref{tab:meld} show the experimental results on the IEMOCAP and MELD data sets. Experimental results show that our method significantly improves the recognition performance of emotion recognition. Specifically, on the MELD dataset, our model improves by 1.40\% and 0.80\% compared to the best-performing baselines W-Acc and W-F1, respectively. Similarly, on the IEMOCAP data set, our model improves by 2.60\% and 2.60\% on W-Acc and W-F1 respectively. The performance improvement may be attributed to the effective extraction of contextual semantic information and efficient integration of underlying data distribution. 

Furthermore, our method is optimal compared with other multi-modal fusion methods in experimental results. The results demonstrate the effectiveness of our model in achieving multi-modal semantic information fusion. We also give W-Acc and W-F1 for each emotion. Specifically, on the IEMOCAP data set, our model's W-Acc is optimal on neutral and frustrated, and W-F1 is optimal on happy, neutral, and frustrated. On the MELD data set, our model's W-Acc is optimal on neutral and frustrated, and W-F1 is optimal on happy, neutral, and frustrated.

We also report the model parameter quantities of the proposed method and the baseline method. The results show that the parameter amount of our model is 1.73M, which is far lower than other methods. The model complexity of other baseline methods is relatively high, but the emotion recognition effect is relatively poor. Experimental results show that the larger the number of parameters, the better the model performance is not necessarily. Experimental results demonstrate that the proposed method is an effective and efficient MERC model.
	
\subsection{Running Time}
In this section, we report the inference time of different baselines and our proposed method on the IEMOCAP and MELD datasets. As shown in Table \ref{tab:running}, the inference time of our method is below 10s, which is much lower than some GCN-based methods and RNN-based methods. Experimental results demonstrate the high efficiency of SSMs.

\begin{table}[htbp]
	\caption{We tested the running time of the proposed method in this paper and other comparative methods on the IEMOCAP and MELD data sets.}
	\label{tab:running}
	\setlength{\tabcolsep}{11mm}{
		\begin{tabular}{@{}l|cc@{}}
			\toprule
			\multirow{2}{*}{Methods} & \multicolumn{2}{c}{Running time (s)} \\ \cline{2-3} 
			& IEMOCAP            & MELD            \\ \midrule
			bc-LSTM      &  8.3       &   10.4   \\
			LMF              & 4.5     & 8.4      \\
			DialogueRNN          &  61.7    & 138.2    \\
			RGAT           & 68.5        &  146.3      \\
			DialogueGCN              & 58.1                 & 127.5             \\
			LR-GCN                   & 87.7                 & 142.3             \\
			DER-GCN                  & 125.5                & 189.7             \\ \hdashline
			Ours                  & 3.5               & 6.6            \\ \bottomrule
	\end{tabular}}
\end{table}

\subsection{Ablation Studies}
\textbf{Ablation studies for PE, BLS, PFM.} As shown in Table \ref{tab:abla}, we found that the performance of the model will decrease after removing PE, which indicates that positional encoding information is quite important for understanding contextual semantic information. Furthermore, without BLS, the performance of the model also degrades. The performance degradation is attributed to the underlying contextual data distribution which is also crucial for emotion prediction. Finally, when the PFM module is removed, the performance of the model drops sharply. Experimental results demonstrate the necessity of multi-modal feature fusion. Different modalities contribute differently to the model's understanding of emotional information, and using label probabilities can guide the model to adaptively learn the weights of different modal features to better integrate multi-modal features.

\begin{figure}
	\centering
	\includegraphics[width=1\linewidth]{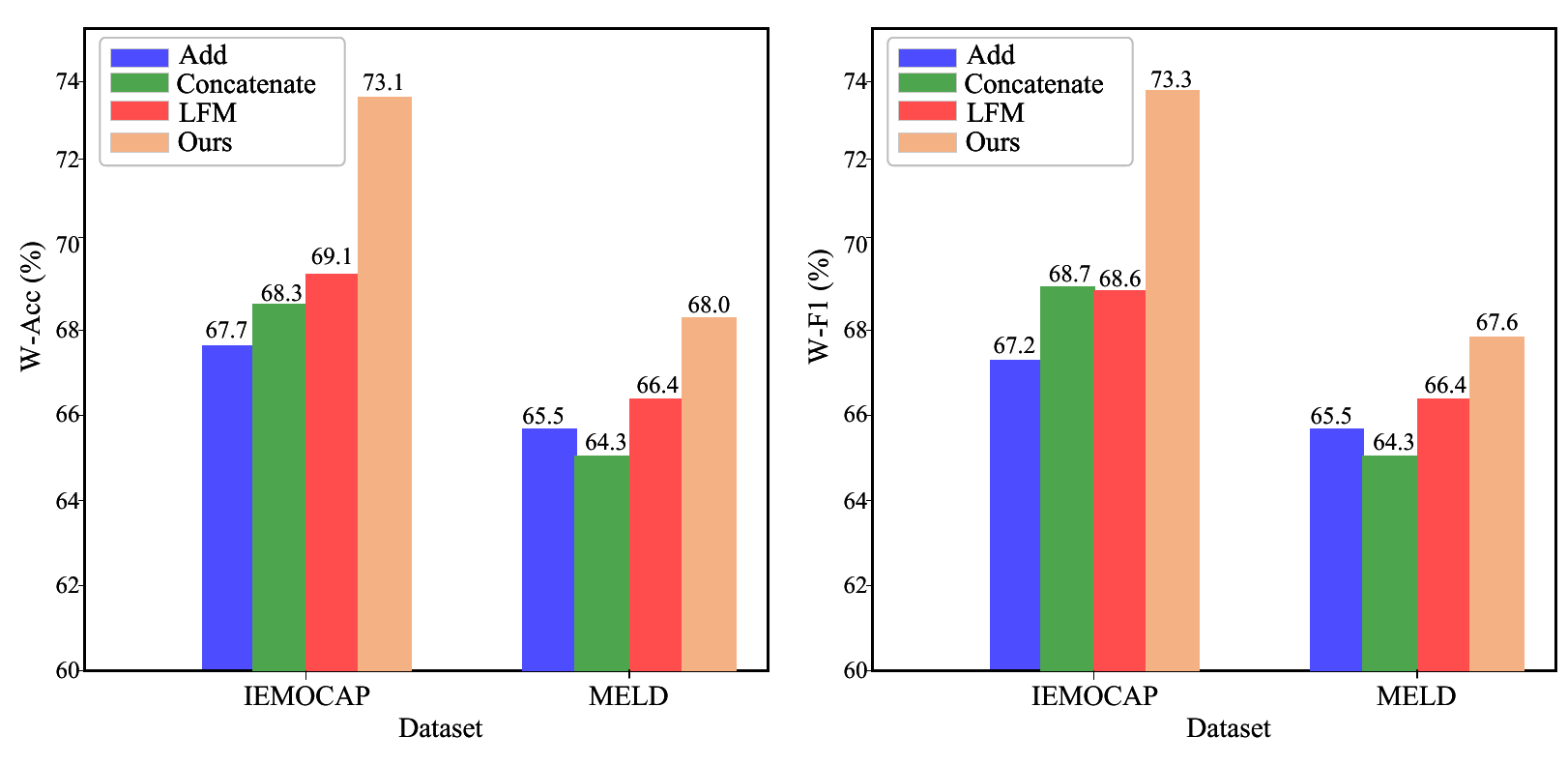}
	\caption{Emotion recognition effects of different fusion methods on the IEMOCAP and MELD datasets. The experimental results are  statistically significant ($t$-test with $p < 0.05$).}
	\label{fig:fusion}
\end{figure}

\begin{figure*}[htbp]
	\centering
	\subfloat[Origin representations]{
		\begin{minipage}[t]{0.24\linewidth}
			\centering
			\includegraphics[width=1.8in]{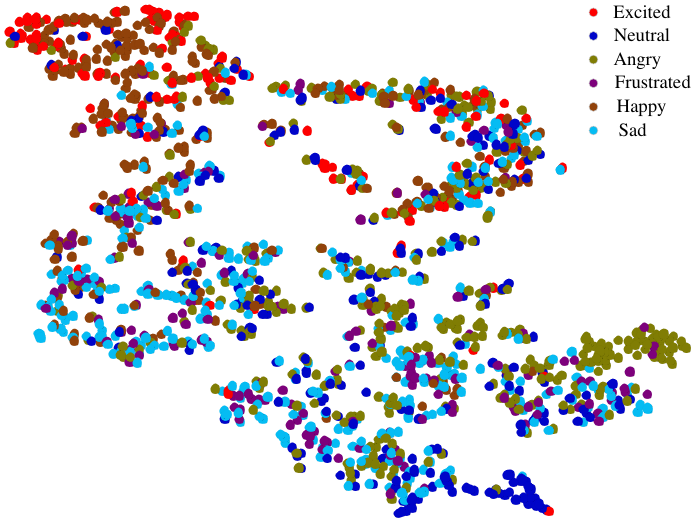}
		\end{minipage}%
	}%
	\subfloat[Learned by our method]{
		\begin{minipage}[t]{0.24\linewidth}
			\centering
			\includegraphics[width=1.8in]{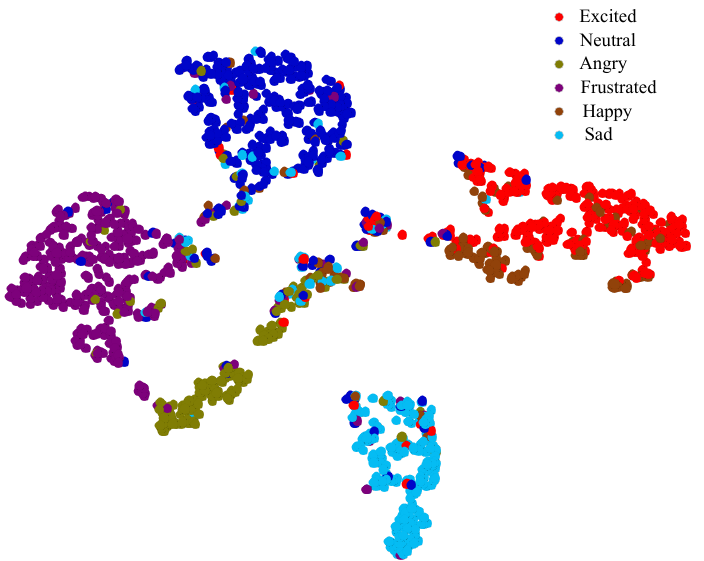}
		\end{minipage}%
	}%
	\subfloat[Origin representations]{
		\begin{minipage}[t]{0.24\linewidth}
			\centering
			\includegraphics[width=1.8in]{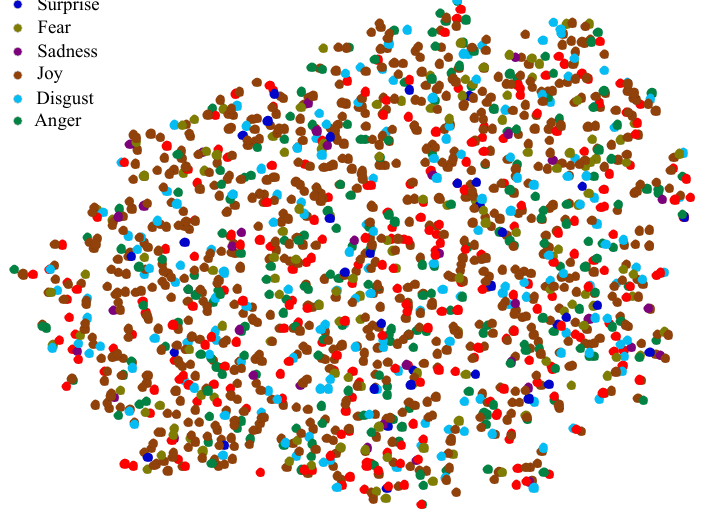}
		\end{minipage}%
	}%
	\subfloat[Learned by our method]{
		\begin{minipage}[t]{0.24\linewidth}
			\centering
			\includegraphics[width=1.8in]{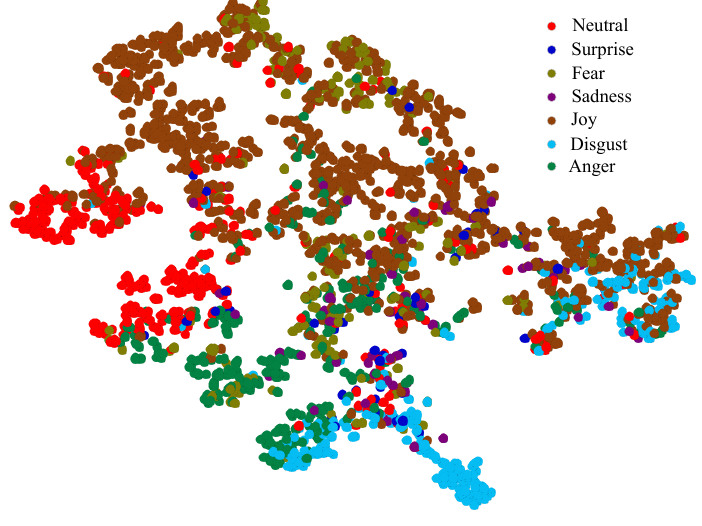}
		\end{minipage}%
	}%
	\centering
	\caption{Visualizing feature embeddings for the multi-modal emotion on the IEMOCAP and meld benchmark dataset. Each dot represents an utterance, and its color represents an emotion. (a) Original features on the IEMOCAP dataset. (b) Features learned by our method on the IEMOCAP dataset. (c) Original features on the MELD dataset. (d) Features learned by our method on the MELD dataset.}
	\label{fig:vis}
\end{figure*}

\textbf{Ablation studies for multi-modal features.} To show the impact of different modal features on experimental results, we conducted ablation experiments to verify the combination of different modal features. From the experimental results in Table IV, it is found that: (1) In the single-modal experimental results of the model, the accuracy of emotion recognition in text mode is far better than the other two modes, indicating that text features play a dominant role in emotion recognition. effect. The emotion recognition effect of video modality features is the worst (2) The emotion recognition effect using bimodal features is better than its own single-modality result. Furthermore, since text features play a dominant role in emotion recognition, this results in the bimodal feature combination with text modality performing better than the combination of acoustic and visual modalities. (3) The emotion recognition effect using three modal features is optimal. Experimental results prove the necessity of fusion of multi-modal features for emotion recognition.

\textbf{Effect of Different Fusion Strategies.} To study the effectiveness of the probability-guided fusion method proposed in this paper, we compare it with some previous multi-modal fusion strategies: (1) Add: multi-modality is implemented by element-wise addition of multi-modal features. Information fusion. (2) Concatenate: directly concatenating multi-modal features. LFM: feature fusion is achieved by introducing a low-rank tensor fusion network.

As shown in Fig. \ref{fig:fusion}, compared with other fusion methods, the probability-guided fusion strategy we proposed has better emotion recognition effects on the two data sets. The results show that the emotion recognition effect of directly adding or concatenating multi-modal features to achieve multi-modal information fusion is relatively poor. The multi-modal information fusion effect of LMF is better than the adding method and the concatenating method. The probabilistic fusion strategy we propose introduces label information to guide the fusion of multi-modal information and further achieves parameter optimization of the model. Interestingly, the fusion effect of the concatenate method on the IEMOCAP data set is better than that of the add method, but the effect on the MELD data set is worse than that of the add method. This may be because the dimensionality of the multi-modal features of the MELD dataset is relatively high, so dimensionality disaster may occur by concatenating multi-modal information. In contrast, our proposed probabilistic guided model effectively uses label information to guide the consistency of multi-modal semantic information.

\subsection{Multi-modal Representation Visualization}
In order to intuitively demonstrate the classification results of our proposed method on the two data sets, we use t-SNE to project the high-dimensional multi-modal feature representation into a two-dimensional space, as shown in Fig. \ref{fig:vis}. The results show that the proposed method is able to effectively separate different emotion categories from each other. However, there are still a small number of samples that are inseparable. In the future, we will consider constructing more stringent target constraints to optimize the distribution of different emotion categories in the feature space, so as to further improve the emotion recognition performance.

\begin{figure}
	\centering
	\includegraphics[width=1\linewidth]{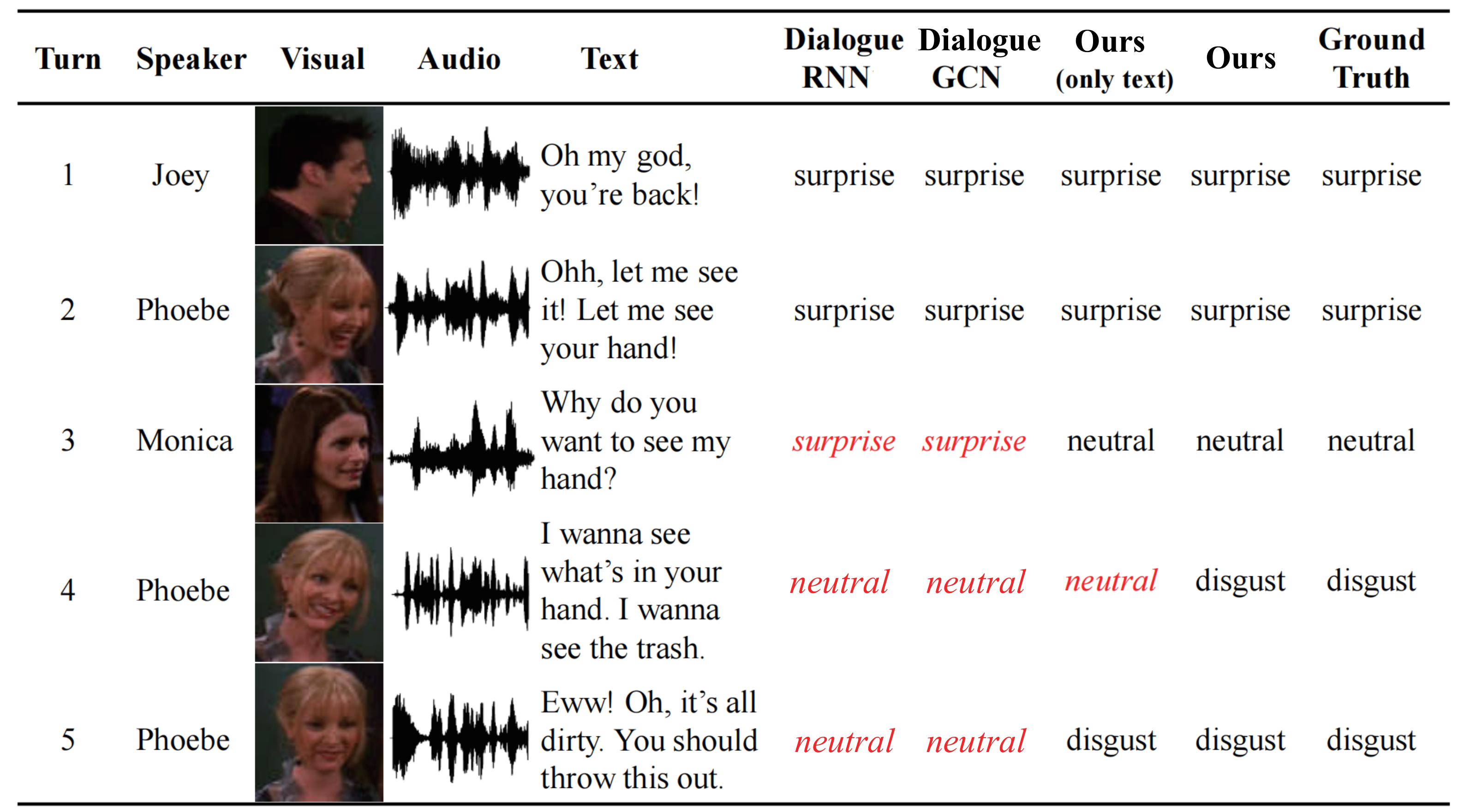}
	\caption{An illustrative example of multi-modal emotion recognition in the MELD dataset. We test the emotion recognition effects of DialogueRNN, DialogueGCN and the proposed method.}
	\label{fig:lizi}
\end{figure}

\subsection{Error Analysis}
As shown in Fig. \ref{fig:lizi}, we test the emotion classification results of DialogueRNN, DialogueGCN and the proposed method on the MELD dataset. In the disgust emotion category, the classification results of DialogueRNN and DialogueGCN are very poor, and they are all misclassified as neutral emotions. When the proposed method only uses text features, the emotion classification effect on the disgust category is unstable, but when multi-modal features are used, it can better classify disgust category emotions.

\section{Conclusions}
In this work, we introduce a novel MERC method that comprehensively considers both feature disentanglement and multi-modal feature fusion. Specifically, during the feature disentanglement, we designed the broad Mamba, which incorporates the SSMs for data-dependent global emotional context modeling, and a broad learning system to explore the potential data distribution in the broad space. Thanks to the proposed bidirectional SSMs, our method can efficiently extract global long-distance contextual semantic information, while only having linear complexity. During the multi-modal feature fusion, we propose an effective probability-guided fusion mechanism to achieve multi-modal contextual feature fusion, which utilizes the predicted label probability of each modal feature as the weight vectors of the modal features. Extensive experiments conducted on two widely used benchmark datasets, IEMOCAP and MELD demonstrate the effectiveness and efficiency of our proposed method.
	
\bibliographystyle{ACM-Reference-Format}
\bibliography{refs}
	
\end{document}